\documentclass[10pt,twocolumn,letterpaper]{article}

\usepackage[pagenumbers]{cvpr} 

\definecolor{cvprblue}{rgb}{0.21,0.49,0.74}
\usepackage[pagebackref,breaklinks,colorlinks,allcolors=cvprblue]{hyperref}

\def\paperID{*****} 
\def\confName{CVPR}
\def\confYear{2026}

\usepackage{algorithm}
\usepackage{algpseudocode}
\usepackage{booktabs}
\usepackage[dvipsnames]{xcolor}
\usepackage{amssymb}
\usepackage{multirow}  
\usepackage{bm}
\usepackage[table]{xcolor}
\usepackage{amsmath}
\usepackage{pifont}
\newcommand{\cmark}{\ding{51}} 
\newcommand{\xmark}{\ding{55}} 

\newcommand{\green}[1]{\textcolor{ForestGreen}{#1}}

\usepackage{tikz}
\usetikzlibrary{calc,backgrounds}

\title{Diagonal-Tiled Mixed-Precision Attention for Efficient Low-Bit MXFP Inference}

\author{
Yifu Ding$^{1,2}$ \quad
Xinhao Zhang$^{4}$ \quad
Jinyang Guo$^{1,3}$\\
$^1$State Key Laboratory of Complex \& Critical Software Environment  (SKLCCSE), Beihang University\\
$^2$School of Computer Science and Engineering, Beihang University\\
$^3$School of Artificial Intelligence, Beihang University\\
$^4$Beijing Jiaotong University\\
{\tt\small yifuding@buaa.edu.cn, 25120417@bjtu.edu.cn, jinyangguo@buaa.edu.cn}
}

\begin{document}
\maketitle

\begin{abstract}
Transformer-based large language models (LLMs) have demonstrated remarkable performance across a wide range of real-world tasks, but their inference cost remains prohibitively high due to the quadratic complexity of attention and the memory bandwidth limitations of high-precision operations. 
In this work, we present a low-bit mixed-precision attention kernel using the microscaling floating-point (MXFP) data format, utilizing the computing capability on next-generation GPU architectures. Our Diagonal-Tiled Mixed-Precision Attention (DMA) incorporates two kinds of low-bit computation at the tiling-level, and is a delicate fused kernel implemented using Triton, exploiting hardware-level parallelism and memory efficiency to enable fast and efficient inference without compromising model performance. 
Extensive empirical evaluations on NVIDIA B200 GPUs show that our kernel maintains generation quality with negligible degradation, and meanwhile achieves significant speedup by kernel fusion. We release our code at \href{https://github.com/yifu-ding/MP-Sparse-Attn}{https://github.com/yifu-ding/MP-Sparse-Attn}. 
\end{abstract}
\section{Introduction}
\label{sec:intro}

The rapid development of large language models (LLMs) has created a growing demand for faster throughput. Inside the transformer architecture, self-attention becomes the main inference bottleneck because its cost scales quadratically with sequence length~\cite{vllm,flashattention,flashattention2,keles2023computational}. To reduce this cost, prior work has explored several directions, including quantization-based compression~\citep{llm-int8}, structured sparsity~\citep{sparsegpt}, and kernel-based approximations such as linear attention~\citep{efficient-attention-survey}. At the same time, modern GPU architectures are adding stronger support for low-precision computation. In particular, NVIDIA Blackwell introduces native support for microscaling number formats, including MXFP8, MXFP4, and NVFP4, achieves lower bitwidth with better quantization quality~\citep{nvidia_blackwell_rtx_2025,nvidia_nvfp4_2025}. These advances make low-precision LLM inference practical in real deployment.

Despite this promising hardware support, effectively leveraging MX formats to accelerate attention in LLMs presents significant challenges. \textbf{\textit{Challenge 1:}} low-bit quantization can lead to severe accuracy degradation. Directly applying 4-bit MXFP formats to attention computation causes substantial quantization error (see \cref{tab:quantization-error}). While recent work~\citep{sageattention3} introduces smoothing techniques to mitigate quantization errors, it requires additional floating-point \texttt{GEMV} operations that reduce kernel throughput. 
\textbf{\textit{Challenge 2:}} unfused operations in quantization process undermine the efficiency of low-bit computation. Our experiment reveals that without kernel fusion, the quantization and format conversion brings redundant memory accesses and kernel launch cost that cannot be underestimated (refer to \cref{tab:not_fused_breakdown}). 

\begin{table}[t]
\centering
\caption{Comparison of representative MXFP data formats, including their block sizes, element formats, and shared scale formats.}
\vspace{-0.05in}
\label{tab:mx_formats}
\resizebox{\linewidth}{!}{
\begin{tabular}{p{1cm} p{1cm} | p{3cm} p{0.5cm} | p{1.2cm}  p{0.5cm}}
\toprule
\multirow{2}{*}{\textbf{Name}} & \textbf{Block} & \multicolumn{2}{c|}{\textbf{Element}} & \multicolumn{2}{c}{\textbf{Scale}} \\ 
& \textbf{Size} &  \textbf{Format} & \textbf{Bits} & \textbf{Format} & \textbf{Bits} \\
\midrule
MXFP8  & 32 & FP8 (E4M3 / E5M2) & 8 & E8M0 & 8  \\
MXFP4  & 32 & FP4 (E2M1)  & 4 & E8M0 & 8 \\
NVFP4  & 16 & FP4 (E2M1)  & 4 & E4M3 & 8 \\
\bottomrule
\end{tabular}
}
\vspace{-0.1in}
\end{table}

In this paper, we propose \textbf{Diagonal-Tiled Mixed-Precision Attention} (DMA), the first attention workflow designed to operate hybrid MX formats. DMA addresses the above challenges via two key techniques: First, we adopt a tiling-level mixed-precision design that partitions the attention matrix into low- and high-precision regions to retain the most salient information for critical tokens in high-precision, while leveraging faster low-precision MX formats elsewhere to ensure the speedup. 
Second, we build a full-stack attention computation workflow into a fused memory-efficient kernel to ensure the end-to-end efficiency. It covers quantization, microscaling transformation, low-bit encoding and packing, and attention computation. This fusion allows fine-grained parallel execution on GPU thread blocks without storing intermediate results, significantly reducing memory pressure and synchronization overhead. 

We evaluate DMA on Longbench dataset. Experimental results show that DMA achieves lossless generation quality compared to full-precision attention baseline. Furthermore, our ablation studies demonstrate the effectiveness of mixed-precision tiling, quantization granularity, and diagonal window design in balancing performance and precision. 

Our main contributions are summarized as follows:
\begin{itemize}
    \item We propose {Diagonal-Tiled Mixed-Precision Attention} (DMA), a new attention workflow that operates with hybrid MX formats for efficient LLM inference.
    
    \item We develop a {fully fused GPU kernel} that integrates quantization, microscaling transformation, and attention computation into one end-to-end workflow, reducing memory access and kernel launch overhead.
    
    \item We conduct extensive experiments and  detailed ablation studies to show the lossless generation quality of our kernel, and offer practical insights into the trade-off between efficiency and accuracy.
\end{itemize}
\section{Related Works}

\subsection{Efficient Attention}

Under the context of large language models (LLMs), the computational overhead of attention scales quadratically with sequence length, making efficient attention a critical research direction. 
FlashAttention~\citep{flashattention}  introduced a highly efficient attention kernel that uses tiling and an online softmax approach to eliminate the need to buffer full attention matrices. 
Quantization-based approaches such as INT‑FlashAttention~\cite{int-flashattention} and SageAttention~\cite{sageattention}  series compress attention operands into low-bit formats. They are plug-and-play quantized attention kernels that achieves 2--3$\times$ speedup over FlashAttention. 
Sparse attention methods like SparseAttention~\cite{zhang2025spargeattention}  limit the number of active token pairs to reduce complexity while preserving important information. As a lightweight sparse attention alternative, TurboAttention~\cite{turboattention} combines head‑wise quantization and sparsity‑aware softmax approximation to deliver 1.2--1.8$\times$ attention speed up and $4\times$ KV cache compression. 

\subsection{MXFP Quantization}
The recent introduction of Microscaling Floating-Point (MXFP) formats is an advancement over traditional Block Floating Point (BFP), designed to improve the numerical efficiency and deployment flexibility of low-precision computations in AI workloads~\citep{rouhani2023microscaling}.
As supported by NVIDIA's latest Blackwell GPU architecture, MXFP4 and MXFP8 allow for significantly higher theoretical throughput, up to {2$\times$ to 4$\times$} compared to FP16 while still maintaining competitive numerical accuracy in mixed-precision settings~\cite{nvidia_blackwell_rtx_2025}.
However, despite these theoretical advantages, systematic and practical software support for MXFP-based attention kernels remains limited, especially for end-to-end pipelines that operate below the standard IEEE precision formats. 
This motivates our first contribution: a full-stack low-bit attention pipeline for MXFP formats (e.g., MXFP8, MXFP4, NVFP4), including efficient quantization and packing, low-bit computation combined with OnlineSoftmax technique. 
\section{Preliminaries}

\subsection{MXFP Data Format}
Microscaling (MX) formats decompose a tensor into low-precision elements and a shared exponent scale per block (typically 32 or 16 elements). This approach allows dynamic range coverage much larger than conventional floating-point formats while significantly reducing storage and compute cost. 
Table~\ref{tab:mx_formats} summarizes the MX formats used in this work. MX formats adopt a shared scale in E8M0 for every 32 elements per block and are named by prefixing ``MX'' to the element data format, i.e., MXFP8 (with FP8 data of E5M2/E4M3) or MXFP4 (FP4 data of E2M1). Since the exponent of FP32 uses 8-bit, the representable range of these formats covers that of FP32, ensuring compatibility with high-dynamic-range input distributions. 
While NVFP4 has FP8 (E4M3) shared scale for every 16 elements in each block. The finer-grained scaling and quantization granularity significantly reduces quantization error compared to MXFP4, enhancing downstream accuracy. 

\subsection{Online Softmax}
\label{sec:flash_attn}
To enable faster attention computation by dividing them into blocks, many efficient attention kernels uses OnlineSoftmax~\cite{flashattention,sageattention} to ensure the equivalence of the results. For example, FlashAttention avoids materializing and storing the full attention matrix by computing the attention output in a tile-wise fashion. It fuses the attention score computation, softmax normalization, and value aggregation into a single memory-efficient kernel, using online row-wise softmax. 

Each attention head is processed in parallel across blocks. $L_Q, L_K$ are the sequence lengths of query and key/value matrices, respectively.  $D$ is the head dimension.
We divide $\mathbf Q$ into tiles of size $B_M$. This results in $\lceil L_Q / B_M \rceil$ thread blocks executing in parallel per batch per head, each responsible for a tile of query $\mathbf Q_i \in \mathbb{R}^{B_M\times D}$. Within each thread block, it iterate over the key/value for $\lceil L_k / B_N \rceil$ tiles.
Let $\mathbf Q_i \in \mathbb{R}^{B_M \times D}$ denote the $i^\text{th}$ query tile, and $\mathbf K_j, \mathbf V_j \in \mathbb{R}^{B_N \times D}$ denote the $j^\text{th}$ key and value tiles respectively. For each $(i,j)$ tile pair, the attention score matrix is computed as:
\begin{equation}
\label{eq:flash_attn}
\mathbf S_{ij} = \mathbf{\tilde Q}_i \times \mathbf{K}_j^\top, \quad \mathbf{\tilde Q}_i = \frac{\mathbf{Q}_i}{\sqrt{D}}.
\end{equation}
It then applies {OnlineSoftmax} across key/value tiles to eliminate the need to store the full matrices of intermediate results. It tracks the running maximum of $\mathbf{S}_i$ in $\bm m_i$ and accumulates the intermediate attention outputs $\mathbf{O}_{i-1}$ across all previous tiles, which is normalized by the ratio of the adjacent maximum values. The accumulated output $\mathbf{O}_i$ and normalization factor $\bm l_i$ are updated incrementally, tile by tile. We finalize the results of the $i^\text{th}$ query tile by $\text{diag}(\bm l_i)^{-1}\mathbf{O}_i$.

\section{Challenges in Low-bit MXFP Attention}
\label{sec:challenge}

\label{sec:acc-degradation-in-low-bit-mxfp}
\begin{figure}
    \centering
    \includegraphics[width=\linewidth]{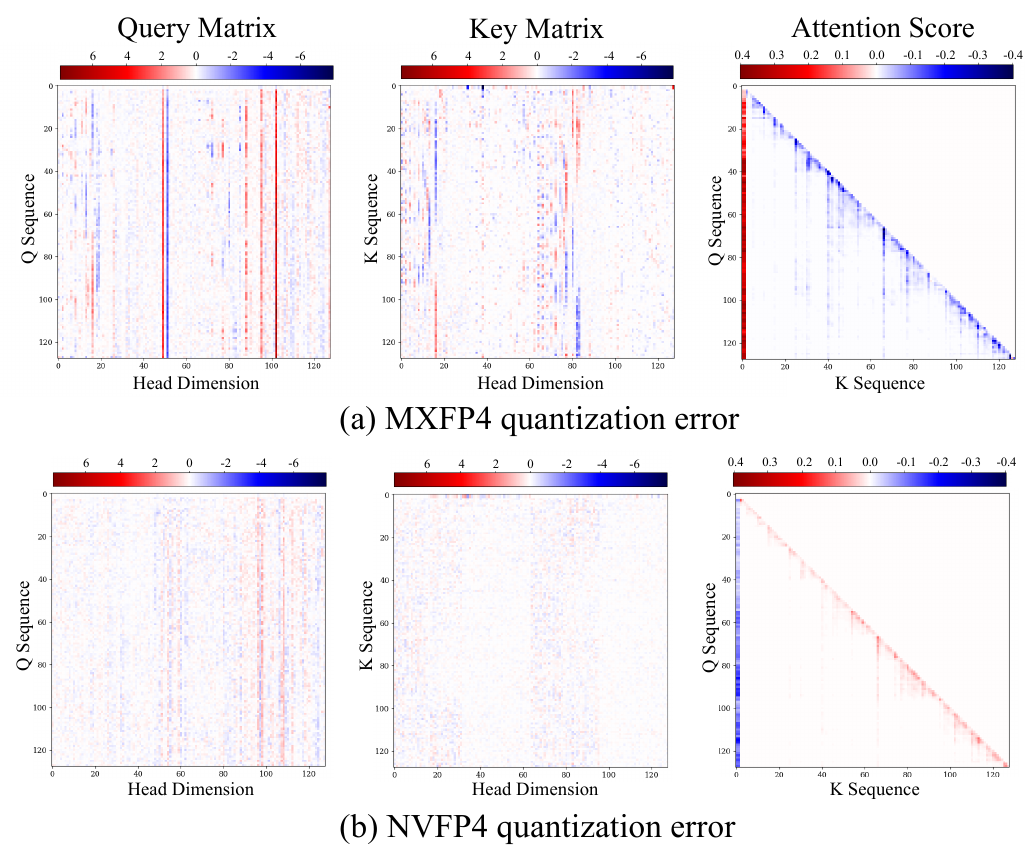}
     \vspace{-0.1in}
    \caption{Visualization of quantization error of MXFP4 and NVFP4 format for query, key and attention score. }
    \label{fig:quant_error}
    \vspace{-0.1in}
\end{figure}

Although full 4-bit quantization can reduce computation and memory cost, directly applying MXFP4 to the $QK^\top$ leads to clear degradation in attention quality. As shown in \cref{fig:quant_error}, the quantization error of MXFP4 is much pronounced than that of NVFP4. 
Table~\ref{tab:quantization-error} confirms this trend quantitatively. Lower-bitwidth MX formats introduce larger quantization errors under all metrics. 
Direct MXFP4 causes a clear drop in attention-score fidelity, with cosine similarity decreasing from 0.988 to 0.714, PSNR from 71.70 to 60.82, and L1 error increasing from 0.246 to 0.924. NVFP4 is notably more stable, and adding token-wise quantization on top of it brings only marginal changes. 

We observe that the error exhibits a clear channel-wise structure in query and key matrices, indicating that some feature dimensions are consistently more sensitive to low-bit quantization than others. However, this channel dimension is exactly the reduction axis in the $QK^\top$ multiplication. Therefore, it is unable to use channel-wise scaling in a fused low-bit attention kernel without introducing substantial implementation complexity. 

\begin{table}[]
    \centering
    \caption{Quantization error of attention score using different data formats. ``+'' means combining with tokenwise quantization. }
    \resizebox{\linewidth}{!}{
    \begin{tabular}{ccccccc}
    \toprule
     \textbf{Format}  & \textbf{Cos Sim} ($\uparrow$) & \textbf{PSNR} ($\uparrow$) & \textbf{L1} ($\downarrow$) & \textbf{RMSE}  ($\downarrow$) \\ \midrule
    MXFP8   & 0.988 & 71.70 & 0.246 & 0.003 \\
    MXFP4   & 0.714 & 60.82 & 0.924 & 0.009 \\
    NVFP4   & 0.982 & 69.37 & 0.309 & 0.003 \\
    NVFP4+  & 0.983 & 69.63 & 0.312 & 0.003 \\ 
    \cellcolor[HTML]{D3D3D3}\textbf{Ours}   & \cellcolor[HTML]{D3D3D3}{\textbf{0.988}} & \cellcolor[HTML]{D3D3D3}{\textbf{71.70}} & \cellcolor[HTML]{D3D3D3}{\textbf{0.248}} & \cellcolor[HTML]{D3D3D3}{\textbf{0.003}} \\
    \bottomrule
    \end{tabular}}
    \label{tab:quantization-error}
\end{table}


\section{Diagonal-Tiled Mixed-Precision Attention}
\label{sec:method}

In this section, we present the overall design of our method, which is built on the typical tiling-style of FlashAttention, but extends it with two components tailored for low-bit MXFP inference: a diagonal-tiled mixed-precision computation scheme to preserve the most sensitive attention regions (\cref{sec:diagonal-tiled}), and a fused quantization kernel to avoid the overhead of separate low-bit pre-processing steps (\cref{sec:kernel_fusion}).

\begin{figure*}[t]
    \centering
    \includegraphics[width=\linewidth]{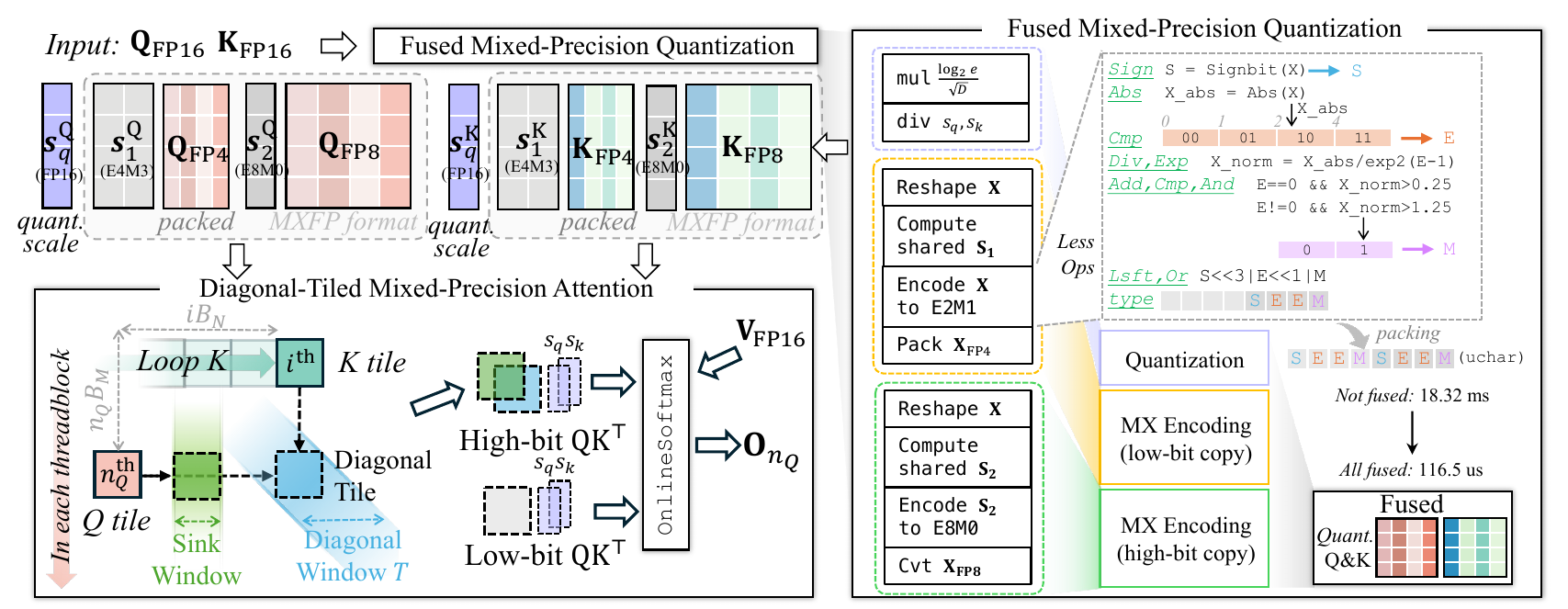}
    \caption{Overview workflow of our Diagonal-Tiled Mixed-Precision Attention. It first applies fused mixed-precision quantization to produce low-bit and high-bit $Q$ and $K$ representations, and then performs diagonal-tiled mixed-precision attention, using higher precision near the diagonal and lower precision elsewhere to balance accuracy and efficiency. }
    \label{fig:overview}
\end{figure*}

\subsection{Overall Workflow}

We follow the tiled execution pattern of FlashAttention to partition attention into sub-tensors for memory-efficient and parallel computation. Built on this workflow, DMA introduces a quantization-aware attention kernel that supports tile-level mixed precision on GPU tensor cores with native MXFP support. 
Meanwhile, we fuse the low-bit pre-processing pipeline into a single Triton implementation. Instead of executing quantization, low-bit encoding, packing, and scale conversion as separate operators, DMA performs them within the kernel, which reduces intermediate memory traffic, kernel launch overhead, and synchronization cost. This fused design is important for maintaining the end-to-end efficiency of mixed-precision attention. The overall framework is illustrated in \cref{fig:overview}. 

\subsection{Diagonal-Tiled Attention Workflow} 
\label{sec:diagonal-tiled}
\begin{algorithm}[t]
\caption{Diagonal-Tiled Mixed-Precision Attention}
\label{algo:diagonal-tile}
\begin{algorithmic}[1]
\Require $\mathbf{\tilde Q}^l \in  \mathbb{R}^{B_M \times D}_\text{FP4}$, $\mathbf{\tilde Q}^h \in\mathbb{R}^{B_M \times D}_\text{FP8}$ are pre-processed with softmax scale. $\mathbf{K}^l \in \mathbb{R}^{B_N \times D}_\text{FP4}$. $\mathbf{K}^h \in \mathbb{R}^{B_N \times D}_\text{FP8}$.
$V \in \mathbb{R}^{B_N \times D}_\text{FP16}$. 
$B_M$, $B_N$: tile sizes in sequence dimension for query and key, respectively. $n_Q$: thread block index. $T$: diagonal window size for high-precision.  

\State Initialize: $\bm m_0 \gets -\infty$, $\bm l_0 \gets 1$, $\mathbf{O}_0 \gets \mathbf{0}$

\noindent \green{Phase 1: Low-precision tiles}
\For{$i\gets0$ \textbf{to} $\lceil (n_Q B_M-T)/B_N \rceil$}
    \State $\mathbf{S}_{i} \gets \mathbf{\tilde Q}_{n_Q}^l\times \mathbf{K}_i^{l\top}$
    \State $\mathbf{O}_{i}, \bm l_i, \bm m_i \gets \text{OnlineSoftmax} (\mathbf{S}_{i}, \mathbf{V}_i, \mathbf{O}_{i-1}, \bm{l}_{i-1}, \bm{m}_{i-1})$
    
\EndFor

\noindent \green{Phase 2: High-precision tiles in diagonal window}
\For{$i \gets \lceil (n_Q B_M-T)/B_N \rceil$ \textbf{to} $\lceil n_Q B_M / B_N \rceil$}
 
$\triangleright$ Skip upper triangular part for \textit{causal attention}
        \State $\mathbf{S}_{i} \gets \mathbf{\tilde Q}_{n_Q}^h\times \mathbf{K}_i^{h\top}$
        \State $\Delta_{i} \gets  \left( n_Q B_M + \mathcal{I}_{B_M} \right) < \left(i B_N +\mathcal{I}_{B_N}\right)$ 
        \State $\mathbf{S}_{i} \gets \mathbf{S}_{i} - \inf \cdot \Delta_{i}$
        
        $\triangleright$ Mask upper triangular of the global attention
        \State $\mathbf{O}_{i}, \bm l_i, \bm m_i \gets \text{OnlineSoftmax}(\mathbf{S}_{i}, \mathbf{V}_i, \mathbf{O}_{i-1}, \bm{l}_{i-1}, \bm{m}_{i-1})$ 

    \EndFor
\State $\mathbf{O}_{n_Q} \gets \text{diag}(\bm l_i)^{-1} \mathbf{O}_i $
\Ensure Causal attention output $\mathbf{O}_{n_Q} \in \mathbb{R}^{B_M \times D}$

\end{algorithmic}
\end{algorithm}

We propose Diagonal-Tiled Mixed-Precision strategy to mitigate the accuracy degradation. As detailed in ~\cref{sec:challenge}, low-bit quantization introduces significant quantization error, and the most influential scores typically concentrate along the diagonal of the attention matrix. Prior solutions, such as SageAttention2~\cite{sageattention2}, compensates for quantization error by performing additional GEMV operations between full-precision $K$ and the mean-pooled query vector, which brings extra full-precision computation overhead. To mitigate this, we propose a mixed-precision strategy in tiling-level that selectively retains higher-precision tiles near diagonal, while aggressively quantizing peripheral regions to ensure the overall throughput. 


Based on the tiling-wise attention paradigm, we introduce a diagonal window with size $T$, determining the amount of tokens for higher-precision computation along the token dimension. For each query tile indexed by $n_Q$, only the last $T$ tokens are computed using higher-precision representations of query and key, while all preceding tiles are aggressively quantized. Taking causal attention as an example, we detail the workflow in Algorithm~\ref{algo:diagonal-tile} in two phases. 

\paragraph{Phase 1.} Starting from the left columns in head dimension. Low-precision attention tiles are computed over the lower triangular region using low-bit quantized query and key matrices. We follow the standard OnlineSoftmax to incrementally accumulate the intermediate output $\mathbf{O}_i$. The first phase terminates after completing the computation for the first $\lceil (n_Q B_M-T)/B_N \rceil$ tokens. 

\paragraph{Phase 2.} Key/value tiles between $\lceil (n_Q B_M-T)/B_N \rceil$ and $\lceil n_Q B_M / B_N \rceil$ will be processed in this phase. To obtain a more precised attention weight, $\mathbf {\tilde Q}^h_{n_Q}, \mathbf K_i^h$ of the high-precision copy will be used.
In causal attention, 
$\Delta_i$ masks scores within the current tile that correspond to the upper triangular region of the global attention matrix, where $\mathcal{I}_{B_M}$ and $\mathcal{I}_{B_N}$ are indices within each tile. It ensures that no query attends to future positions globally. 
Final outputs are normalized by the accumulated softmax scaling factors $\bm l_i$.

This diagonal-tiled execution maintains the performance and memory benefits of quantized attention while preserving the numerical precision of the attention mechanism to keep the generation quality. 

\paragraph{Compatibility with Non-Causal Attention}

While Algorithm~\ref{algo:diagonal-tile} illustrates the workflow of causal attention, our method is also compatible with non-causal attention. 
In the non-causal attention workflow, the low-bit computation covers both the lower and upper triangular regions of the attention matrix, excluding the diagonal window. 

The structure of the workflow remains similar to that of the causal case, but the iteration range of Phase 1 differs: the tile index $i$ spans two disjoint intervals. Specifically, tiles from $1$ to $\lceil (n_Q B_M - T/2) / B_N \rceil$ corresponds to the lower triangular region, and tiles from $\lceil (n_Q B_M + T/2) / B_N \rceil$ to $\lceil n_K B_N / B_N \rceil$ covers the upper triangular. 
During this phase, the attention weight $\mathbf S_i$ is computed using the low-precision copies of the query and key matrices. 
In Phase 2, tokens located within the diagonal window $[\lceil (n_Q B_M-T/2)/B_N \rceil, \lceil (n_Q B_M + T/2 )/B_N \rceil]$ are reprocessed using high-precision copies of the inputs. The results are then accumulated into $\mathbf{O}_i$ computed in Phase 1. Once all key/value tiles have been iterated over, the final attention output is calculated by $\text{diag}(\bm l_i)^{-1}\mathbf O_i$. 

This partitioning is valid due to the equivalence under column-wise transformations in matrix multiplication. Specifically, the dot-product attention is a linear operation with respect to the key-value axis, and the output remains mathematically correct as long as all tiles of the attention matrix are covered and accumulated appropriately. Hence, splitting the computation into lower, upper triangular and diagonal regions does not compromise correctness, provided the transformations are applied consistently across tiles.

\subsection{Dual MXFP Quantization Kernel Fusion}
\label{sec:kernel_fusion}

In this section, we describe the fused mixed-precision quantization kernel used in DMA. The goal is to convert the input tensor into both low-bit and high-bit MXFP representations within a single fused pipeline, so that the subsequent mixed-precision attention kernel can directly consume the quantized outputs without launching another kernels. The overall procedure is summarized in Algorithm~\ref{algo:mixed-precision-quant}. 

\begin{algorithm}[t]
\caption{Fused Mixed-Precision Quantization}
\label{algo:mixed-precision-quant}
\begin{algorithmic}[1]
\Require $\mathbf{X} \in \mathbb{R}^{B \times D}$: input FP16 tensor. $B$: size of a tile in sequence dimension, $D$: size of head dimension. $V_1$, $V_2$: sizes of a block for low-/high-precision MXFP formats. 
   $l_1, l_2,u_1,u_2$: lower/upper bound for both elements formats. 
   $e^{\max}$: the exponent of the largest normal number of the element format. 

   \noindent \green{Step 1: Pre-process softmax scale}
   \If {\text{is Query}} 
   \State $\mathbf{X}_\text{sm-scaled} \gets \mathbf{X} \cdot \frac{\log_2 e}{\sqrt{D}}$ 
   \EndIf

   \noindent \green{Step 2: Compute the quantization scale}
   \State $\mathbf{S}_q \gets \max_{D}(|\mathbf{X}_\text{sm-scaled}|) / (448\times6)$
   \State $\mathbf{X}_\text{scaled} \gets \mathbf{X}_\text{sm-scaled} / \mathbf{S}_q$
   
   \noindent \green{Step 3: Compute shared scale for low precision format}
   \State $\mathbf{X}_\text{scaled}' \gets \mathbf{X}_\text{scaled} \quad \triangleright \text{reshape to } [B, D//V_1, V_1]$
   \State $\mathbf{S}_{\text{FP4}} \gets \max_D(|\mathbf{X}_\text{scaled}^{'}|) / u_1$ 
   \State $\mathbf{X}_{\text{clamped}} \gets \text{clamp}(\mathbf{X}_\text{scaled}' / \mathbf{S}_{\text{FP4}}, l_1, u_1)$
   
   \noindent \green{Step 4: Encode $\mathbf{X}_{\text{clamped}}$ to E2M1}
   \State $\mathbf{X}_\text{FP4} \gets $ Algorithm~\ref{algo:float4-quantization}~($\mathbf{X}_{\text{clamped}}$)
   
   \noindent \green{Step 5: Pack two FP4 into one UINT8 along $D$}
   \If{$\text{pack along last dimension}$} 
       \State $\mathbf{X}_\text{FP4}' \gets \mathbf{X}_\text{FP4} \quad \triangleright \text{reshape to } [B, (D + 1) // 2, 2] $
       \State $\mathbf{L, H} \gets \mathbf{X}_\text{FP4}'[:, :, 0], \mathbf{X}_\text{FP4}'[:, :, 1]$
       \State $\mathbf{X}_{\text{packed}} \gets (\mathbf{H} \ll 4) \;|\; \mathbf{L}$
   \EndIf
   
   \noindent \green{Step 6: Compute shared scale for high precision format}
   \State $\mathbf{X}_\text{scaled}^{''} \gets \mathbf{X}_\text{scaled} \quad \triangleright \text{reshape to } [B, D//V_2, V_2]$
   \State $\mathbf{S}_\text{shared} \gets \lfloor \log_2(\max_D|\mathbf{X}_\text{scaled}^{''}|) \rfloor - e^{\max}$
   \State $\mathbf{X}_\text{FP8} \gets \text{clamp}(\mathbf{X}_{scaled}^{''} / 2^{\mathbf{S}_{\text{shared}}}, l_2, u_2)$
   
   \noindent \green{Step 7: Convert shared scale into E8M0}
   \State $\mathbf{S}_\text{FP8} \gets \text{clamp}(\mathbf{S}_\text{shared} + 127, 0, 254)$
   \Ensure Packed FP4 tensor $\mathbf{X}_{\text{packed}}$, FP8 tensor $\mathbf{X}_{\text{FP8}}$, shared scale of NVFP4 $\mathbf{S}_{\text{FP4}}$, shared scale of MXFP8 $\mathbf{S}_{\text{FP8}}$, quantization scale $\mathbf{S}_q$
\end{algorithmic}
\end{algorithm}

\paragraph{Pre-process softmax scale (Step 1).} 
Before quantization, we first apply the standard softmax scaling factor to the query tensor. Specifically, when the input is $Q$, we multiply it by $\log_2 e / \sqrt{D}$, where $D$ is the head dimension. This incorporates the softmax normalization factor into the quantized computation in advance, so that the subsequent $QK^\top$ accumulation is already aligned with the scaled attention score formulation. Since our kernel computes the matrix product in base-2 arithmetic, folding this factor into the query in advance avoids an extra scaling step after accumulation and simplifies the fused implementation.

\paragraph{Quantization scale (Step 2).} 
We fuse quantization with MXFP number format conversion at the kernel level. Following the two-level scaling strategy in SageAttention3, we note that the shared scale in NVFP4 uses the FP8 (E4M3) format, which ranges from $[-448,448]$, while each element is represented in FP4, with a dynamic range of  $[-6,6]$. Therefore, the representable range of NVFP4 is the product of these two bounds. To fully utilize the available dynamic range and accommodate potential outliers without resorting to clipping, we compute the quantization scale in Step 2 (line 4) of Algorithm~\ref{algo:mixed-precision-quant}, scaling the original tensor into this representable range prior to quantization. 

\paragraph{Compute shared scale (Step 3 and Step 6).} We first reshape the scaled input tensor along the packing dimension into a new matrix, where each row groups the $V_1$ or $V_2$ elements that share a common scale in the MXFP format. For NVFP4, the shared scale is computed directly as the absolute maximum of each group. For MXFP4 and MXFP8 formats, however, the scale is represented in integer E8M0 format. 
To maximize the representable range, we need to normalize the input exponent so that the largest exponent in the data aligns with the maximum representable exponent (denoted as $e^{\text{max}}$) of the low-bit element format. Consequently, the shared exponent $\mathbf{S}_\text{shared}$ stores this offset from the input's exponent to $e^{\text{max}}$. 
In E5M2, the maximum exponent $e^{\text{max}}$ is 15 (i.e., $(11110)_2=30$ with a bias of 15, excluding reserved patterns with all 1 bits to represent infinite and NaN). In E4M3, $e^{\text{max}} = 8$ (i.e., $(1111)_2=15$ with a bias of 7). Notably, E4M3 does not strictly follow IEEE-754, its normal numbers allows exponent bits set to all 1. The division of exponent shared scale allows full utilization of the exponent range within the limited bit budget of MXFP format.

\begin{algorithm}[t]
\caption{Encode FP16 tensor into E2M1 format}
\label{algo:float4-quantization}
\begin{algorithmic}[1]
\Require $\mathbf{X}\in \mathbb{R}_\text{FP16}^{B \times D} \in [-6.0, 6.0]$: clamped FP16 tensor. 

\noindent \green{Step 4.1: Extract signbit}
\State $\mathbf{S} \gets \text{sign}(\mathbf{X})$

\noindent \green{Step 4.2: Compute 2-bit exponent}
\State $\mathbf{X}_\text{abs} \gets |\mathbf{X}|$
\State $\mathbf{E} \gets \mathbb{I}[\mathbf{X}_\text{abs} \geq 1] + \mathbb{I}[\mathbf{X}_\text{abs} \geq 2] + \mathbb{I}[\mathbf{X}_\text{abs} \geq 4]$

\noindent \green{Step 4.3: Compute 1-bit mantissa}
\State $bias \gets 1$
\State $\mathbf{X}_{\text{norm}} \gets \mathbf{X}_\text{abs} / 2^{\mathbf{E} - bias}$
\State $\mathbf{M} \gets \mathbb{I}[ \mathbf{E} = 0 ] \cdot \mathbb{I}[ \mathbf{X}_{\text{norm}} > 0.25 ] + \mathbb{I}[ \mathbf{E} \ne 0 ] \cdot \mathbb{I}[ \mathbf{X}_{\text{norm}} > 1.25 ]$ 

\noindent \green{Step 4.4: Assemble quantized FP4 integer}
\State $\mathbf{X}_{\text{FP4}} \gets (\mathbf{S} \ll 3) \;|\; (\mathbf{E} \ll 1) \;|\; \mathbf{M}$

\Ensure Quantized tensor $\mathbf{X}_{\text{FP4}} \in \mathbb{R}_\text{FP4}^{B \times D}$ 
\end{algorithmic}
\end{algorithm}

\paragraph{Encoding of FP4 format (Step 4).}  
Algorithm~\ref{algo:float4-quantization} shows the steps to quantize an FP16 tensor into E2M1 format (1-bit sign, 2-bit exponent, 1-bit mantissa). We begin by extracting the sign bit (Step 4.1). The 2-bit exponent is determined by thresholding the absolute value of the input against \{1, 2, 4\}, yielding exponent values in \{0, 1, 2, 3\}, which can be represented by 2 bits (Step 4.2). We employ the Kronecker delta function $\mathbb{I}[\cdot]$ in line 3 and 6, which returns $1$ if the condition is true and $0$ otherwise. In this way, subnormal values are assigned $E=0$. To compute mantissas (Step 4.3), We normalize the input by the implied scale factor and assign the mantissa bit to 1 if $\mathbf{X}_\text{norm}$ exceeds 0.25 (the midpoint of 0 and 0.5). For normal values ($E\neq0$), the mantissa is set to 1 if $\mathbf{X}_\text{norm}$ exceeds the midpoint of the two representable values under that exponent. For example, when $E=3$, E2M1 can represent 4 and 6, which correspond to normalized values of 1 and 1.5, making 1.25 the comparison threshold for mantissa. 
To implement \texttt{roundTiesToEven} according to the standard, we prefer rounding to even mantissas (i.e., $M=0$) in tie-breaking scenarios. For example, for input value 5, we prefer rounding to 4 to have mantissa bit as 0, rather than 6. So the comparison to midpoint threshold should be greater but not equal. In final, step 4 consrtucts the 4-bit representation using bitwise shift and OR operations.

\paragraph{Packing (Step 5).} We encode two FP4 values into a single byte, assigning the value with the higher index to the most significant 4 bits and the other to the least significant 4 bits. This compact representation improves memory bandwidth utilization.

\paragraph{Scale Conversion into E8M0 (Step 7).} 
The shared scale for MXFP8 should be in E8M0 (unsigned 8-bit integer) format according to the official document of MXFP computing. Therefore, we add 127 before clamping it into 0 to 254, ensuring it lies within the valid exponent range of E8. 

\section{Experiments}

We conduct a comprehensive evaluation of the proposed DMA operator in terms of both accuracy and efficiency. We report results across a range of tasks and datasets to assess the overall performance. In addition, we perform detailed ablation studies to analyze the impact of key design choices, including numerical precision, quantization granularity, diagonal window size, and kernel fusion strategies.

\subsection{Settings}

\paragraph{Models and Datasets.} We evaluate our method using LLaMA-3.1-8B and LLaMA-3.2-3B~\cite{llama3}. Performance is measured on LongBench~\cite{longbench}, which focuses on long-context understanding with sequence lengths ranging from 2.5K to 30K tokens. We report results across various subtasks to assess general long-context capabilities. For all reported metrics, higher values indicate better performance.

\paragraph{Implementation.} We implement DMA in Triton~\cite{triton}. We compare our method with the native attention kernel, which refers to the SDPA kernel originally supported in \texttt{PyTorch} and computed in BF16 format~\cite{sdpa}. All experiments are conducted on a single NVIDIA B200 GPU.

\subsection{Accuracy}

\begin{table}[ht]
\centering
\caption{Comparison of attention implementations on  LLaMA3.2-3B  and  LLaMA3.1-8B. ``Native'' refers to SDPA implementation supported by PyTorch.}
\label{tab:attn_comparison_dtype}
\begin{tabular}{lrrrr}
\toprule
 & \multicolumn{2}{c}{\textbf{LLaMA3.1-8B}} & \multicolumn{2}{c}{\textbf{LLaMA3.2-3B}} \\ 
\textbf{Task} & \textbf{Native} & \textbf{Ours} & \textbf{Native} & \textbf{Ours} \\
\midrule
2wikimqa & 39.15 & 32.54 & 30.75 & 36.33 \\
dureader & 28.40 & 32.93 & 28.11 & 33.04 \\
gov\_report & 33.87 & 34.89 & 32.26 & 33.04 \\
hotpotqa & 49.89 & 50.49 & 46.18 & 48.59 \\
lcc & 49.46 & 56.96 & 42.98 & 45.54 \\
lsht & 41.50 & 44.50 & 26.75 & 30.75 \\
multi\_news & 26.16 & 27.07 & 25.56 & 21.78 \\
multifieldqa\_en & 52.26 & 52.65 & 46.84 & 47.87 \\
multifieldqa\_zh & 57.57 & 58.81 & 48.06 & 53.62 \\
musique & 25.82 & 26.81 & 20.35 & 25.77 \\
narrativeqa & 25.28 & 26.70 & 19.76 & 25.44 \\
passage\_count & 4.02 & 6.22 & 3.40 & 1.50 \\
passage\_retrieval\_en & 99.00 & 96.00 & 80.00 & 37.00 \\
passage\_retrieval\_zh & 92.14 & 85.50 & 8.25 & 10.50 \\
qasper & 42.64 & 43.05 & 32.13 & 39.62 \\
qmsum & 23.47 & 24.67 & 22.52 & 23.63 \\
repobench-p & 42.23 & 54.00 & 44.14 & 50.15 \\
samsum & 36.60 & 43.99 & 34.79 & 42.74 \\
trec & 65.00 & 71.50 & 62.50 & 69.50 \\
triviaqa & 77.01 & 87.89 & 83.68 & 88.13 \\
vcsum & 14.80 & 17.92 & 13.65 & 16.69 \\
\midrule
\textbf{Avg.} & 44.11 & \cellcolor[HTML]{D3D3D3}\textbf{46.43} & 35.84 & \cellcolor[HTML]{D3D3D3}\textbf{37.20} \\
\bottomrule
\end{tabular}
\end{table}

\paragraph{Performance on LongBench}

We evaluate our method on LongBench to assess long-context language understanding performance. Table~\ref{tab:attn_comparison_dtype} shows that our method improves the average score over the native attention baseline on both LLaMA3.1-8B and LLaMA3.2-3B. 
Specifically, for LLaMA3.1-8B, the average score improves from 44.11 to 46.43. For LLaMA3.2-3B, the average score improves from 35.84 to 37.20. The gains are broad across many tasks, including \texttt{repobench-p}, \texttt{samsum}, \texttt{trec}, and \texttt{triviaqa}. In particular, \texttt{repobench-p} improves from 42.23 to 54.00 on LLaMA3.1-8B and from 44.14 to 50.15 on LLaMA3.2-3B. 
The average results across both models indicate that our method preserves, and in most cases improves, long-context accuracy compared with the native implementation.

\subsection{Efficiency}

Table~\ref{tab:format_latency} reports the latency breakdown of different formats and block-scale configurations. We compare our implementation with several fixed-format baselines, including MXFP4, NVFP4, and MXFP8. For each setting, we report the attention time, quantization overhead, and total runtime. 

Among all evaluated settings, our configuration with diagonal and sink sizes set to 128 achieves the lowest total latency, at 7.776 ms. This is lower than MXFP4 (12.980 ms), NVFP4 (13.404 ms), and MXFP8 (16.771 ms). In particular, the main reduction comes from the attention kernel time, which is 7.110 ms in our 128/128 setting, compared with 12.491 ms, 12.941 ms, and 16.480 ms for the three baselines, respectively. 
We also evaluate a larger block-scale configuration with diagonal and sink sizes set to 256. In this case, the total latency increases to 15.720 ms. Compared with the 128/128 setting, this suggests that a larger block size is less efficient in our current implementation. 

\begin{table}[h]
\centering
\caption{Latency breakdown of different block-scale types and configurations. ``MP Size'' denotes the mixed-precision block size used for the higher-bit diagonal and sink blocks.} 
\resizebox{0.95\linewidth}{!}{
\begin{tabular}{lccrrr}
\toprule
\textbf{Format} & \textbf{MP Size} & \textbf{Attn (ms)} & \textbf{Quant (ms)} & \textbf{Total (ms)} \\
\midrule
MXFP4   & –    & 12.491    & 0.242        & 12.980     \\
NVFP4   & –    & 12.941    & 0.204        & 13.404     \\
MXFP8   & –    & 16.480    & 0.044        & 16.771     \\  \midrule
\cellcolor[HTML]{D3D3D3}\textbf{Ours}   & \cellcolor[HTML]{D3D3D3}\textbf{128}& \cellcolor[HTML]{D3D3D3}\textbf{7.110}     & \cellcolor[HTML]{D3D3D3}\textbf{0.382}        & \cellcolor[HTML]{D3D3D3}\textbf{7.776}      \\
Ours    & 256 & 15.056    & 0.382        & 15.720     \\
\bottomrule
\end{tabular}
}
\label{tab:format_latency}
\end{table}



\subsection{Ablation Study}

\begin{table}[t]
\centering
\caption{Similarity metrics under different token numbers for diagonal and sink windows. }
\label{tab:similarity-metrics}
\vspace{-0.05in}
\resizebox{\linewidth}{!}{
\begin{tabular}{ccrrrrr}
\toprule
\textbf{Diag.} & \textbf{Sink} & \textbf{Bit$_\text{high}$ (\%)} & \textbf{Cos Sim} $\uparrow$ & \textbf{Rel. L1 $\downarrow$} & \textbf{RMSE $\downarrow$} & \textbf{PSNR} $\uparrow$  \\ \midrule
- & - & 0.0  & 0.778 & 0.620 & 0.065 & 43.715 \\
- & - & 100.0 & 0.819 & 0.547 & 0.059 & 44.568 \\
\midrule
0 & 128 & 1.15  & 0.781 & 0.780 & 0.072 & 42.817 \\
128 & 0 & 1.15  & 0.782 & 0.644 & 0.066 & 43.635 \\
\textbf{128} & \textbf{128} & \textbf{2.30}  & \textbf{0.822} & \textbf{0.539} & \textbf{0.059} & \textbf{44.657} \\
512 & 512 & 9.22  & 0.826 & 0.542 & 0.058 & 44.731 \\
2048 & 2048 & 36.87 & 0.852 & 0.521 & 0.054 & 45.352 \\

\bottomrule
\end{tabular}}
\end{table}

\paragraph{Mixed-Precision Block Tile Sizes. }
We conduct ablation experiments to evaluate the impact of different block tile sizes on the similarity between the quantized attention matrix and its full-precision counterpart. As shown in \cref{tab:similarity-metrics}, we report the similarity before and after quantization to show the representation errors under varying diagonal and sink tile configurations, including Cosine Similarity, Relative L1 Distance, Root Mean Square Error (RMSE), and Peak Signal-to-Noise Ratio (PSNR). The Bit$_\mathrm{high}\%$ column denotes the percentage of values in the attention matrix that compute in high-precision. 

We observe that increasing the tile size from 128 to 512 improves similarity metrics marginally. But it leads to a significant degradation in throughput due to reduced parallelism and slower computation as shown in \cref{tab:format_latency}. 
Based on this trade-off, we use the 128/128 configuration as the default setting in the following experiments. 

\paragraph{Kernel Fusion}
\begin{table}[t]
\centering
\caption{Ablation study of kernel fusion components to analyze their effect on throughput (TOPS) and latency. \textbf{Encode}: encoding FP16 to FP4/FP8 MX format. \textbf{Pack}: packing two FP4 values to one UINT8. \textbf{Scale Cvt.}: converting microscaling scalar to E8M0 format. \textbf{MP}: fusing quantization for mixed bitwidth to a single kernel. }
\label{tab:kernel-fusion-components-ablation}
\vspace{-0.1in}
\resizebox{\linewidth}{!}{
\begin{tabular}{ccccrr}
\toprule
\textbf{Encode} & \textbf{Pack} & \textbf{Scale Cvt.} & \textbf{MP} & \textbf{$L=2\text{k}$ ($\mu$s)} & \textbf{$L=8\text{k}$ ($\mu$s)}  \\
\midrule
\xmark & \xmark & \xmark & \xmark & 7262.41 & 22628.96 \\
\cmark & \xmark & \xmark & \xmark & 802.90 & 1113.77 \\
\cmark & \cmark & \xmark & \xmark & 740.64 & 942.67 \\
\cmark & \cmark & \cmark & \xmark & 179.97 & 299.69 \\
\cellcolor[HTML]{D3D3D3}\textbf{\cmark} & \cellcolor[HTML]{D3D3D3}\textbf{\cmark} & \cellcolor[HTML]{D3D3D3}\textbf{\cmark} & \cellcolor[HTML]{D3D3D3}\textbf{\cmark} & \cellcolor[HTML]{D3D3D3}\textbf{97.87} & \cellcolor[HTML]{D3D3D3}\textbf{282.46} \\
\bottomrule
\end{tabular}}
\end{table}

\begin{table}[t]
\centering
\caption{Latency breakdown of the non-fused MX encoding pipeline and our fused implementation.}
\vspace{-0.1in}
\label{tab:not_fused_breakdown}
\resizebox{0.88\linewidth}{!}{
\begin{tabular}{lrr}
\toprule
\textbf{Operator} & \textbf{Time} & \textbf{Time (\%)} \\
\midrule
\textbf{Not fused} & 18.320 ms & -- \\
\textbf{ - Element encoding} & 17.742 ms & 100.0\% \\
\qquad MinOps & 2.110 ms & 11.89\% \\
\qquad ArgMinOps & 2.054 ms & 11.58\% \\
\qquad Direct\_Copy & 1.256 ms & 7.08\% \\
\qquad CompareEq & 890.18 us & 5.02\% \\
\qquad AddOps & 810.56 us & 4.57\% \\
\qquad MulFunctor & 755.10 us & 4.26\% \\
\qquad Memcpy / Memset & 206.85 us & 1.17\% \\
\textbf{ - Element packing} & 81.22 us & 100.0\% \\
\qquad BitwiseOr & 23.74 us & 29.24\% \\
\qquad lshift & 22.63 us & 27.86\% \\
\textbf{ - Scalar Convert} & 521.47 us & 100.0\% \\
\qquad IndexOps & 35.75 us & 6.85\% \\
\qquad DeviceSelectSweep & 25.73 us & 4.93\% \\
\qquad Write\_Indices & 20.42 us & 3.91\% \\
\qquad Direct\_Copy & 17.12 us & 3.28\% \\
\qquad Memcpy & 13.57 us & 2.60\% \\
\midrule
\cellcolor[HTML]{D3D3D3}\textbf{Kernel Fusion (Ours)} &  \cellcolor[HTML]{D3D3D3}\textbf{116.480 us} &  \cellcolor[HTML]{D3D3D3}{--} \\
\bottomrule
\end{tabular}}
\end{table}

We evaluate the contribution of each kernel fusion component in \cref{tab:kernel-fusion-components-ablation}. 
The fully unfused baseline, which executes all quantization-related steps separately, incurs a latency of 7262.41~$\mu$s for sequence length $L=2$k and 22628.96~$\mu$s for $L=8$k. 
When only in-kernel FP16-to-MX encoding is enabled, the latency is reduced to 802.90~$\mu$s for $L=2$k and 1113.77~$\mu$s for $L=8$k.
After further enabling FP4 packing, which packs two FP4 values into one \texttt{uint8}, the latency is further reduced to 740.64~$\mu$s for $L=2$k and 942.67~$\mu$s for $L=8$k.
When scale conversion of microscaling factors to the \texttt{e8m0} format is also fused into the kernel, the latency drops substantially to 179.97~$\mu$s for $L=2$k and 299.69~$\mu$s for $L=8$k. 
Finally, after integrating mixed-precision quantization into a single fused kernel, denoted as \textbf{MP}, the latency is further reduced to 97.87~$\mu$s for $L=2$k and 282.46~$\mu$s for $L=8$k, achieving the best overall performance. Compared with the fully unfused baseline, this corresponds to a 74.2$\times$ speedup for $L=2$k and an 80.1$\times$ speedup for $L=8$k. 

Overall, these results demonstrate that each fusion component contributes to latency reduction, and a fully fused kernel is critical for achieving an efficient mixed-precision attention.




\begin{table}[t]
\centering
\caption{Performance comparison under different quantization granularities. Latency are reported in milliseconds.}
\label{tab:granularity_latency}
\resizebox{\linewidth}{!}{
\begin{tabular}{lrrcccc}
\toprule
\textbf{Granu.} & \textbf{Latency $\downarrow$} & \textbf{Cos Sim $\uparrow$} & \textbf{Rel. L1 $\downarrow$} & \textbf{RMSE $\downarrow$} & \textbf{PSNR $\uparrow$} \\
\midrule
Per-Tensor & 6.276 ms & 0.732 & 0.560 & 0.067 & 43.479 \\
Per-Block & 6.366 ms & 0.736 & 0.558 & 0.067 & 43.531 \\
{Per-Token} & 7.131 ms &{0.822} &{0.539} &{0.059} &{44.657} \\
\bottomrule
\end{tabular}
}
\end{table}

\paragraph{Quantization Granularity. }
We further investigate the impact of different quantization granularities on both latency and kernel output precision. As shown in \cref{tab:granularity_latency}, we evaluate four granularity settings: Per-Tensor, Per-Block, and Per-Token. Latency is measured using the 5 warmups and average of 10 runs. 
The results show that finer granularity, such as Per-Token, achieves the highest output similarity before and after quantization, as indicated by higher Cosine Similarity (0.822), lower RMSE (0.059), and higher PSNR (44.657), but it also incurs the highest latency (7.131 ms). In contrast, Per-Tensor and Per-Block offer lower latency but exhibit larger quantization errors. Therefore, the choice of quantization granularity depends on the task requirement, specifically whether latency or output quality is prioritized.

\section{Conclusion}

In this paper, we present DMA, a diagonal-tiled mixed-precision attention method for low-bit MXFP attention computation for efficient LLM inference. DMA improves the accuracy of low-bit attention by preserving more sensitive diagonal regions in higher precision, while keeping the remaining regions in efficient low-bit formats. Combined with a fused dual-MXFP quantization kernel, our design reduces quantization preprocessing overhead and maintains practical efficiency. Experimental results show that DMA provides a better balance between accuracy and latency.

\section*{Limitation}

Our current study is limited to a small set of model sizes and benchmark settings, and the evaluation is mainly focused on long-context text workloads. In particular, we mainly validate DMA on text-based tasks and do not extend the experiments on vision or vision-language settings. In addition, the mixed-precision tiling policy is not validated on extremely long sequence lengths, diverse hardware settings, or other model architectures and attention variants. Extensive experiments and validations are important directions for our future work. 

\newpage

\appendix



{
    \small
    \bibliographystyle{ieeenat_fullname}
    \bibliography{main}
}


\end{document}